%% file: main.tex
\documentclass[10pt,twocolumn,letterpaper]{article}

\usepackage{cvpr}              
\input{preamble}
\definecolor{cvprblue}{rgb}{0.21,0.49,0.74}
\usepackage[pagebackref,breaklinks,colorlinks,allcolors=cvprblue]{hyperref}
\usepackage{booktabs}
\usepackage{multirow}
\usepackage{xcolor}
\usepackage{float} 
\usepackage[accsupp]{axessibility}
\def\paperID{11} 
\def\confName{CVPR}
\def\confYear{2026}

\title{Rethinking Transfer Learning for Industrial Inspection: DINOv3 vs. ImageNet Pretraining Across RGB and X-ray Tasks}
\author{
Mehdi Gharbage$^{1,2}$ \quad Céline Teulière$^{2}$ \quad Pierre Bouges$^{1}$ \quad Thierry Chateau$^{2}$\\[0.5em]
$^{1}$Michelin Tyres Manufacturer \\
$^{2}$Université Clermont Auvergne, CNRS, Institut Pascal, Clermont-Ferrand, France\\
}

\begin{document}
\maketitle
\input{sec/0_abstract}    
\input{sec/1_intro}
\input{sec/2_related_works}
\input{sec/3_methodology}
\input{sec/4_experimental_settings}
\input{sec/5_results_and_analysis}

\input{sec/6_discussions}
\input{sec/7_acknowledgments}

{
    \small
    \bibliographystyle{ieeenat_fullname}
    \bibliography{main_v2}
}


\end{document}

%% file: sec/0_abstract.tex
\begin{abstract}
Vision foundation models pretrained on web-scale data have recently shown strong transfer capabilities on many downstream tasks, but their effectiveness for industrial visual inspection remains unclear. Industrial data differ substantially from web-data and often require fine-grained dense prediction, raising the question of whether modern self-supervised pretraining can improve over the conventional transfer-learning paradigm based on supervised ImageNet initialization. In this work, we compare ConvNeXt backbones pretrained with supervised ImageNet classification or DINOv3 distillation, and relate them to the conventional ResNet-50 baseline. We evaluate semantic segmentation, instance segmentation, and object detection across four downstream datasets spanning RGB surface-defect inspection and X-ray defect detection. We further study both frozen and fully finetuned adaptation regimes. Our results show that DINOv3 offers no clear advantage in frozen transfer, but provides a stronger initialization after full finetuning on RGB tasks, yielding faster convergence and better final performance. Under X-ray modality shift, however, supervised ImageNet pretraining remains more effective in both frozen and finetuned settings. Overall, our findings suggest that modern vision foundation models are promising for supervised RGB industrial inspection, but their transferability is strongly conditioned by downstream adaptation and target modality.
\end{abstract}

%% file: sec/1_intro.tex
\section{Introduction}
\label{sec:intro}
\begin{figure}[t]
    \centering
    \includegraphics[width=\columnwidth]{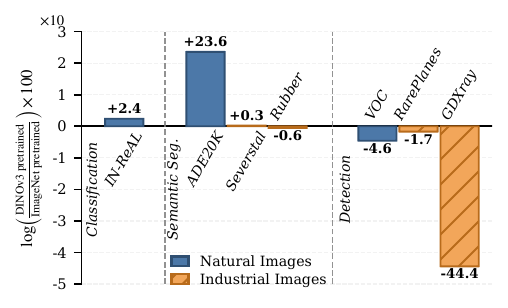}
    \caption{Comparison of DINOv3 distillation and ImageNet supervised pre-training on convolutional backbones ConvNeXts across downstream image classification, semantic segmentation, and object detection under a linear probing setting with frozen backbones. Bars report the relative performance gap between the two pre-training methods: positive values indicate an advantage for DINOv3 and negative values indicate an advantage for ImageNet supervised pre-training. Web-scale benchmark results are taken from prior work for image classification and ADE20K semantic segmentation \cite{simeoniDINOv32025a}, while Pascal VOC, Severstal, RarePlanes, and GDXray results are obtained in our probing setup.}
    \label{fig:pretraining_delta}
\end{figure}

Visual inspection is a critical component of industrial quality control, where the objective is often to detect, localize, and segment small defects under challenging imaging conditions. Unlike natural-image benchmarks \cite{russakovskyImageNetLargeScale2015}, industrial inspection data typically exhibit strong domain-specific characteristics, including limited annotated data, severe foreground/background imbalance, fine-grained local patterns, and large appearance variability across products and acquisition conditions. These difficulties become even more pronounced when moving beyond conventional RGB imagery to modalities such as X-ray, where the visual statistics differ substantially from those of web-scale natural images. As a result, designing transferable visual representations for industrial inspection remains a central challenge.

For many years, the dominant transfer-learning paradigm in computer vision has relied on supervised ImageNet pretraining \cite{heRethinkingImageNetPreTraining2019}, which has served as the standard initialization strategy for downstream recognition and localization tasks. In industrial inspection, this approach remains widely used because it offers a practical solution when task-specific annotations are scarce \cite{akcayAutomaticThreatDetection2022,akcayTransferLearningUsing2016,fergusonAutomaticLocalizationCasting2017}. However, recent progress in vision foundation models \cite{oquabDINOv2LearningRobust2023,simeoniDINOv32025a,raviSAM2Segment2024c,heMaskedAutoencodersAre2022,radfordLearningTransferableVisual2021a} has shifted the focus toward large-scale self-supervised pretraining on web data, with the promise of learning more generic and reusable visual representations. In particular, modern dense-oriented models such as DINOv3 \cite{simeoniDINOv32025a} suggest that large-capacity architectures trained at scale may provide stronger features not only for global recognition, but also for structured prediction tasks such as segmentation and detection.

Whether this promise extends to industrial inspection remains unclear. On the one hand, industrial tasks require spatially precise representations and often involve visual patterns that differ significantly from those seen in natural-image data. On the other hand, if foundation-model features transfer well despite this distribution shift, they could redefine the standard initialization strategy for industrial vision systems. This question is especially important in supervised industrial settings, where the goal is not simply to detect anomalies relative to normal samples, but to learn class-aware and localization-aware predictors for semantic segmentation, instance segmentation, and object detection.

In this work, we study this question through a controlled comparison between conventional and modern transfer-learning paradigms. We take supervised ImageNet pretraining as the conventional baseline and compare it against DINOv3 distillation on the modern convolutional backbones ConvNexts \cite{liuConvNet2020s2022}. More specifically, we evaluate ConvNeXt-T backbones pretrained either with supervised ImageNet classification or with DINOv3 distillation, and we relate these results to a strong ResNet-50 baseline. Our study spans three supervised industrial visual tasks : semantic segmentation, instance segmentation and object detection across four downstream datasets, including both RGB surface inspection and X-ray defect detection. We further analyze transfer under different adaptation regimes, from frozen backbones to full finetuning, in order to determine whether DINOv3 representations can be used directly or whether their benefit emerges mainly through downstream adaptation.

Our results show that the answer is nuanced. On RGB industrial datasets, DINOv3 does not provide a clear advantage when the backbone is frozen, but becomes the strongest initialization once the backbone is fully finetuned, leading to faster convergence and better final performance for semantic segmentation and RGB instance-level localization. In contrast, this benefit does not extend to X-ray imagery, where supervised ImageNet pretraining remains superior in both frozen and fully finetuned regimes. These findings suggest that the value of modern vision foundation models for industrial inspection lies less in direct universal transfer than in their ability to provide strong finetuning priors when the target data remain sufficiently close to natural-image statistics.

Our contributions are : 
\begin{itemize}
    \item We provide a controlled study of transfer learning for supervised industrial inspection across semantic segmentation, instance segmentation, and object detection.
    \item We compare conventional supervised ImageNet pretraining with DINOv3 distillation on convolutional backbones under both frozen and finetuned adaptation regimes.
    \item We show that the benefit of large-scale self-supervised pretraining is strongly dependent on both the downstream task and the target modality, with clear gains on RGB tasks but limited transfer under X-ray modality shift.
\end{itemize}

%% file: sec/2_related_works.tex
\section{Related Works}
\label{sec:related_works}
\subsection{Vision Foundation Models for Dense Prediction.}
Over the past few years, vision foundation models have emerged as a major direction in computer vision, driven by the premise that large-capacity architectures trained on web-scale image collections can learn generic visual representations that transfer effectively across a broad range of downstream tasks. This trend naturally extends the long-standing paradigm of transfer learning, in which supervised pre-training on ImageNet \cite{russakovskyImageNetLargeScale2015} became the standard initialization strategy for many vision models. In particular, ImageNet pre-training has been widely adopted for downstream recognition and localization tasks, often improving optimization and data efficiency when target annotations are limited \cite{heRethinkingImageNetPreTraining2019}, as it enables models to acquire generic low- and mid-level visual cues such as edges, textures, and object parts. However, the scale and diversity of supervised classification data remain inherently limited, which motivated the rise of self-supervised learning as a way to exploit much larger unlabeled image collections. Early self-supervised approaches, including contrastive methods \cite{chenSimpleFrameworkContrastive2020a,caronUnsupervisedLearningVisual2020} and joint-embedding methods \cite{grillBootstrapYourOwn2020a,chenImprovedBaselinesMomentum2020,caronEmergingPropertiesSelfSupervised2021a,assranSelfSupervisedLearningImages2023a}, introduced pretext objectives that enabled representation learning without human supervision. Yet these objectives were primarily optimized for global image-level representations rather than dense visual features. More recent studies \cite{el-noubyAreLargescaleDatasets2021,wangDenseContrastiveLearning2021a} have shown that global pretext tasks are not always well aligned with dense prediction, and that strong global performance does not necessarily translate into strong dense representations \cite{simeoniDINOv32025a}. This observation led to the development of training strategies that more explicitly promote dense visual learning \cite{wangDenseContrastiveLearning2021a,oquabDINOv2LearningRobust2023,heMaskedAutoencodersAre2022,simeoniDINOv32025a,zhouImageBERTPretraining2021}. In particular, masked image modeling approaches \cite{heMaskedAutoencodersAre2022,zhouImageBERTPretraining2021} made dense representation learning a central design objective, while contrastive and joint-embedding methods \cite{wangDenseContrastiveLearning2021a,oquabDINOv2LearningRobust2023,simeoniDINOv32025a} were progressively adapted to better support dense visual features. Among these models, DINOv3 \cite{simeoniDINOv32025a} is especially relevant to our study because it is explicitly designed to preserve and improve dense representations at scale. By adapting the DINO framework to dense prediction and releasing both Vision Transformer and ConvNeXt backbones distilled from a large-scale teacher, DINOv3 provides a particularly suitable foundation for studying transfer to supervised industrial segmentation and detection under strong domain shift.

\subsection{Supervised Transfer Learning for Industrial Inspection.}
Supervised transfer learning has long been a practical and widely adopted paradigm in industrial inspection, where obtaining large-scale defect annotations is often costly and application-specific. Prior to the rise of vision foundation models, the standard approach consisted of initializing convolutional backbones with supervised ImageNet pre-training \cite{russakovskyImageNetLargeScale2015} and fine-tuning them for downstream tasks such as defect classification, object detection, and segmentation. This strategy was particularly attractive in low-data industrial settings because pre-trained models improve optimization and reduce overfitting compared with training from scratch \cite{heRethinkingImageNetPreTraining2019}. Benchmark efforts such as VISION reflect the continued importance of this supervised setting by providing industrial datasets with pixel-level and instance-level annotations for detection and segmentation \cite{baiVISIONDatasetsBenchmark2023}, while recent surveys explicitly distinguish such closed-set supervised formulations from open-set and anomaly-detection paradigms \cite{chengComprehensiveSurveyRealworld2026}. The same transfer-learning recipe has also been explored in X-ray imagery, where convolutional networks pre-trained on natural images have been adapted to object recognition and detection despite the substantial modality gap \cite{akcayTransferLearningUsing2016,hassanDetectingProhibitedItems2020}. More recent studies further show that supervised industrial segmentation remains challenging in low-data regimes, motivating stronger transferable representations beyond conventional ImageNet-initialized CNNs \cite{liuExploringFewshotDefect2025}.
\subsection{Adapting Foundation Models to Industrial Inspection.}
Recent work has begun to explore how foundation models can be adapted to industrial visual inspection, but much of this literature has focused on anomaly detection rather than supervised defect recognition and localization. In particular, several approaches build upon general-purpose vision or vision-language foundation models through task-specific adaptation strategies such as prompt engineering, prompt tuning, or parameter-efficient fine-tuning. Representative examples include CLIP-based anomaly detection methods that learn static and dynamic prompts for industrial zero-shot transfer \cite{caoAdaCLIPAdaptingCLIP2025a}, as well as SAM-based approaches that introduce dedicated tuning modules and adapters to improve anomaly segmentation under industrial domain shift \cite{yangPromptableAnomalySegmentation2025a}. More broadly, recent reviews of industrial foundation models highlight prompt learning, fine-tuning, and domain adaptation as central mechanisms for bringing general-purpose foundation models into manufacturing applications \cite{zhaoIndustrialFoundationModels2025}. While these methods demonstrate that foundation models can be made effective in industrial contexts, they typically do so through additional adaptation methods, auxiliary tuning, or domain-specific refinement, rather than by directly assessing whether the pretrained visual representations themselves are already sufficient for transfer.

A notable exception is AnomalyDINO \cite{dammAnomalyDINOBoostingPatchbased2025}, which investigates whether the frozen visual features of DINOv2 can be directly exploited for industrial anomaly detection without any additional fine-tuning or meta-learning. Its strong results suggest that foundation-model features can already provide a powerful basis for patch-level anomaly scoring. However, this line of work remains tied to the anomaly-detection setting, where the objective is to identify deviations from normality rather than to solve supervised defect detection or segmentation with semantic labels. As a result, such methods do not fully address the requirements of supervised industrial inspection, where models must produce class-aware and spatially precise dense predictions across tasks such as semantic segmentation, instance segmentation, and object detection. Related studies on few-shot defect segmentation further suggest that vision foundation models hold significant promise in industrial settings, but also emphasize the need for careful task adaptation and evaluation beyond conventional anomaly-detection benchmarks \cite{liuExploringFewshotDefect2025}. These observations motivate our focus on a complementary question: whether recent dense-oriented vision foundation models can transfer effectively to supervised industrial inspection tasks under strong domain and modality shift, and how much adaptation is actually required.

%% file: sec/3_methodology.tex
\section{Methodology}
\label{sec:methodology}
\subsection{Study Objective}
The objective of this work is to evaluate whether large-scale self-supervised pretraining on web data can provide a stronger transfer-learning paradigm for supervised industrial visual inspection than the conventional supervised ImageNet pretraining commonly used in practice. In particular, we study whether DINOv3-distilled ConvNets backbones improve transfer to industrial tasks requiring spatially precise predictions, namely semantic segmentation, instance segmentation, and object detection. Beyond overall performance, our study also examines whether representations learned from large-scale natural-image data remain effective under the strong distribution shifts that characterize industrial imagery, including more challenging image modalities such as X-ray data. Through this evaluation, we aim to determine whether DINOv3 pretraining can constitute a new practical standard for transfer learning in supervised industrial defect detection.
\subsection{Comparative Study}
To answer this question, we adopt a controlled comparative study rather than introducing a new task-specific method. As a conventional baseline, we consider a ResNet-50 backbone initialized with supervised ImageNet pretraining, which remains a standard transfer-learning recipe for industrial vision. We then evaluate stronger convolutional backbones based on ConvNeXt \cite{liuConvNet2020s2022} under two different pretraining settings: supervised ImageNet classification pretraining and DINOv3 distillation. This design allows us to reduce the bias induced by architectural differences and to focus more directly on the impact of the pretraining paradigm itself.

Our evaluation is conducted on three supervised industrial inspection tasks: semantic segmentation, instance segmentation, and object detection. For each setting, we study two adaptation regimes: frozen and fully adapted backbones. Applying the same adaptation protocol across pretrained backbones enables a fair comparison with the conventional ResNet baseline and allows us to assess how easily each representation can be transferred to industrial tasks.
\subsection{ConvNets Backbones}
We focus our evaluation on convolutional networks (\textit{ConvNets}) for two main reasons. First, ConvNets remain a strong and widely adopted baseline in industrial inspection, where dense prediction tasks often require accurate localization of small defects, repetitive textures, and fine structural patterns. Their inductive properties, such as locality and translation equivariance \cite{liuConvNet2020s2022,wangTheoreticalAnalysisInductive2023}, are well suited to such scenarios, and their feature hierarchies can be easily integrated with standard decoders for segmentation and detection. Second, restricting the main comparison to convolutional backbones allows us to better isolate the effect of pretraining from the effect of architecture. In particular, comparing ConvNeXt backbones pretrained with supervised ImageNet objectives or with DINOv3 distillation provides a more controlled setting for evaluating whether large-scale self-supervised pretraining brings consistent benefits to industrial inspection. While Vision Transformers \cite{dosovitskiyImageWorth16x162020} may also offer strong transfer capabilities, especially at scale, we leave their role to future analysis in order to keep the present study centered on the impact of pretraining.

%% file: sec/4_experimental_settings.tex
\section{Experimental Settings}
\label{sec:experimental_settings}
\subsection{Architecture Details}
To evaluate the transferability of pretrained convolutional backbones across different supervised industrial vision tasks, we instantiate each backbone within standard dense prediction frameworks. For semantic segmentation, we use Mask2Former \cite{chengMaskedAttentionMaskTransformer2022} as the decoder architecture. For instance segmentation, we use Mask R-CNN \cite{heMaskRCNN2017a}, and for object detection, we use Faster R-CNN \cite{girshickFastRCNN2015}. 

Our study focuses on the effect of pretraining rather than architectural redesign. Therefore, the downstream architectures are kept unchanged across experiments, except for the conventional ResNet-50 baseline. Following \cite{heRethinkingImageNetPreTraining2019}, we replace Batch Normalization with Group Normalization in ResNet-50 in order to ensure stable optimization in detection and segmentation settings. No additional modification is introduced to the backbone or decoder architectures.
\begin{table}[t]
    \centering
    \footnotesize
    \setlength{\tabcolsep}{3pt}
    \caption{Overview of industrial datasets used in this study. $^\dagger$ For GDXray, we split train and validation sets based on series : validation set contains series 3,32,45,54,61 and 87, the rest is for training, except the series 55.}
    \label{tab:datasets}
    \begin{tabular}{lcccc}
        \toprule
        & Severstal & Rubber Rings & GDXray & RarePlanes \\
        \midrule
        Task       & Sem. seg. & Sem. seg. & Obj. Det. & Inst. Seg. \\
        Modality   & RGB       & RGB       & X-ray & RGB \\
        Classes    & 4         & 9        & 1    & 3 \\
        Train      & 5332        & 158        & 706    & 4474 \\
        Val        & 1334        & 39        & 356    & 1113 \\
        Split      & Rand. & Stratified Rand. & Custom$^\dagger$ & Official \\
        Test short edge & 384 & 800 & 800 & 800 \\
        \bottomrule
    \end{tabular}
\end{table}

\subsection{Datasets}
We evaluate our study on four datasets covering complementary industrial inspection settings and task formulations. Together, these datasets allow us to assess transfer across semantic segmentation, instance segmentation, and object detection, while also spanning different appearance statistics and image modalities. A summary of the datasets, including class counts, splits, and preprocessing details, is provided in Table~\ref{tab:datasets}.
\subsubsection{Semantic Segmentation Datasets}

\textbf{Severstal.} We use the Severstal dataset \cite{SeverstalSteelDefect} to evaluate semantic segmentation of surface defects on steel surfaces. This dataset is representative of industrial surface inspection, where defects appear as localized texture and structure alterations on largely homogeneous backgrounds.

\textbf{Rubber Rings.} We further evaluate semantic segmentation on a real-world rubber ring defect dataset \cite{liuExploringFewshotDefect2025}. Compared with Severstal, this dataset is more oriented towards industrial quality control, where the goal is to identify defects on manufactured parts. Plus this dataset contains less training samples, which makes it more challenging and more representative of real-world industrial settings.
\subsubsection{Detection and Localization Datasets}
\textbf{RarePlanes.} For instance segmentation and localization, we use RarePlanes \cite{shermeyerRarePlanesSyntheticData2021} on real aerial images. Although this dataset is not strictly an industrial inspection benchmark, it shares several properties with challenging inspection scenarios, including small objects, multi-scale structure, and significant appearance variability over large backgrounds. We therefore use it as a proxy setting to evaluate how pretrained backbones transfer to fine-grained instance-level localization tasks different from natural images datasets.

\textbf{GDXray Castings.} To study transfer under a stronger modality shift, we use the castings subset of GDXray \cite{meryGDXrayDatabaseXray2015}, which consists of X-ray images with defect annotations. This dataset is particularly important for our work because it enables us to evaluate whether representations learned from large-scale natural-image pretraining remain effective when transferred to industrial X-ray inspection. It therefore serves as a key benchmark for assessing the robustness of foundation-model features beyond conventional RGB imagery.

\begin{table*}[t]
    \centering
    \caption{Comparison of backbone pretraining strategies and finetuning regimes across four downstream evaluation datasets. We report mIoU for the semantic segmentation datasets Severstal and Rubber Rings, mask mAP@[0.5:0.95] for RarePlanes, and box mAP@50 for GDXray. All results are obtained on the validation splits used for each dataset and are reported at the final training iteration. In all cases, the downstream task head is optimized on the target dataset, while the backbone is either frozen or fully finetuned depending on the adaptation setting. All experiments are done at the fixed random seed 42.}
    \small
    \setlength{\tabcolsep}{5.5pt}
    \renewcommand{\arraystretch}{1.12}
    \begin{tabular}{l l l l c c c c}
        \toprule
        & \multicolumn{2}{c}{Pretraining} & & \multicolumn{4}{c}{Downstream dataset} \\
        \cmidrule(lr){2-3} \cmidrule(lr){5-8}
            & 
            & \multicolumn{2}{c}{\textcolor{gray}{Decoder Arch.}}
            & \multicolumn{2}{c}{\textcolor{gray}{Mask2Former}} 
            & \textcolor{gray}{Mask R-CNN}
            & \textcolor{gray}{Faster R-CNN} \\
        \cmidrule(lr){5-6}
        Backbone & Method & Dataset & Finetuning & Severstal & Rubber Rings & RarePlanes & GDXray \\
        \midrule
        ResNet-50
            & Supervised cls. & ImageNet-1k & Full
            & 63.28 & 73.87 & 78.39 & 24.42 \\
        \midrule
        \multirow{6}{*}{ConvNeXt-T}
            & \multirow{3}{*}{Supervised cls.}
            & \multirow{3}{*}{ImageNet-1k}
            & Frozen  & 62.04 & 73.25 & 72.89 & 21.32 \\
            &  &  & Full    & 62.97 & 73.26 & 82.88 & \textbf{29.74} \\
        \cmidrule(lr){2-8}
            & \multirow{3}{*}{DINOv3 distill.}
            & \multirow{3}{*}{LVD-1689M}
            & Frozen  & 62.40 & 72.32 & 70.36 & 7.88 \\
            &  &  & Full    & \textbf{64.01} & \textbf{75.60} & \textbf{84.50} & 27.84 \\
        \bottomrule
    \end{tabular}
    \label{tab:comparison_pretraining}
\end{table*}

\subsection{Training Protocol}
Our goal is not to optimize each dataset independently through extensive hyperparameter search, but rather to evaluate the transferability of different pretraining strategies under a controlled and consistent protocol. Accordingly, we keep the training procedure as close as possible to the default recipes of the considered downstream architectures, and we introduce some modifications when they are necessary as detailed in the next paragraph. This design allows performance differences to be attributed primarily to the pretrained representations rather than to benchmark-specific optimization.

\textbf{Training recipe.} For semantic segmentation, both Severstal and Rubber Rings exhibit a strong foreground/background imbalance, with defect regions occupying only a small fraction of the image area. To reduce the number of crops containing only background, we use a defect-aware cropping strategy during training: with probability $p=0.7$, the crop is constrained to include at least part of an annotated defect region. For object detection on GDXray Castings, we adapt the anchor-box configuration to better match the small defect instances present in the dataset as in \cite{fergusonAutomaticLocalizationCasting2017}. Aside from these task-specific adjustments, all other settings follow the default configurations in Detectron2 \cite{wu2019detectron2}.

\textbf{Scheduling.} We keep the learning-rate scheduling strategy fixed within each task family. For semantic segmentation with Mask2Former, we use the default schedule of the original implementation for all experiments. For detection and instance segmentation, we follow the scheduling strategy of \cite{heRethinkingImageNetPreTraining2019}, consisting of a linear warm-up phase followed by multi-step learning-rate decay. We use a $3\times$ schedule for RarePlanes and a longer $6\times$ schedule for GDXray Castings.

\textbf{Hyperparameters.} To preserve a fair comparison, we do not perform dataset-specific hyperparameter tuning. Instead, hyperparameters are kept fixed across experiments within each architecture family. For semantic segmentation, we retain the default Mask2Former configuration. For detection and instance segmentation, we follow the recipe of \cite{heRethinkingImageNetPreTraining2019} when the backbone is ResNet-50, and we use the standard ConvNeXt hyperparameter setting \cite{liuConvNet2020s2022} when the backbone is ConvNeXt. This controlled protocol is intended to ensure that observed differences mainly reflect the quality of the pretrained features.

All experiments are implemented using the Detectron2 framework \cite{wu2019detectron2}, which provides a unified training and evaluation environment for semantic segmentation, instance segmentation, and object detection.

%% file: sec/5_results_and_analysis.tex
\begin{figure}[ht]
  \centering
   \includegraphics[width=1.0\linewidth]{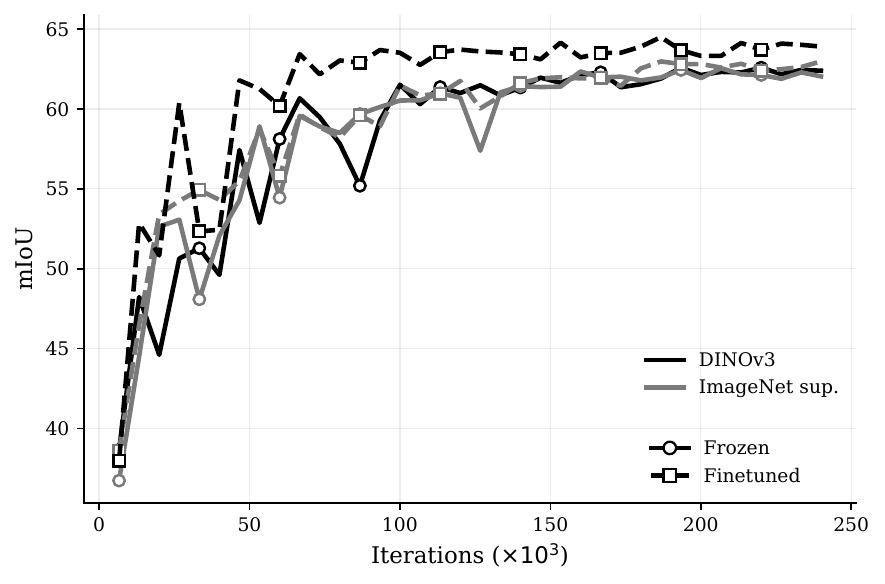}
   \caption{Learning curves for mIoU on the Severstal validation set using Mask2Former with ConvNeXt-T bakbones pretrained with supervised ImageNet classification or DINOv3 distillation.}
   \label{fig:sem_seg_results}
\end{figure}
\section{Results and Analysis}
\label{sec:results_and_analysis}
\begin{figure}[ht]
  \centering
   \includegraphics[width=1.0\linewidth]{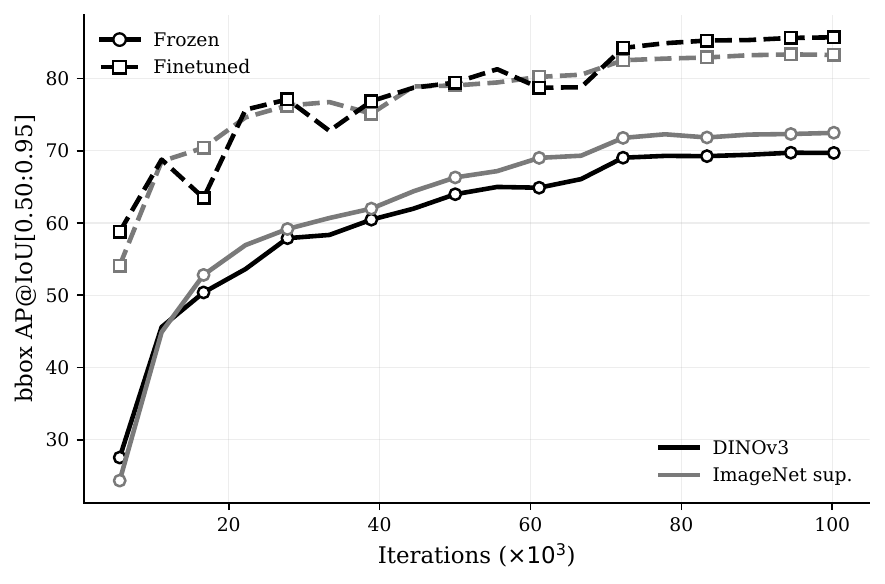}
   \caption{Learning curves for bounding box mAP on the RarePlanes validation set using Mask R-CNN with ConvNeXt-T bakbones pretrained with supervised ImageNet classification or DINOv3 distillation.}
   \label{fig:rareplanes_results}
\end{figure}

\begin{figure}[ht]
  \centering
   \includegraphics[width=1.0\linewidth]{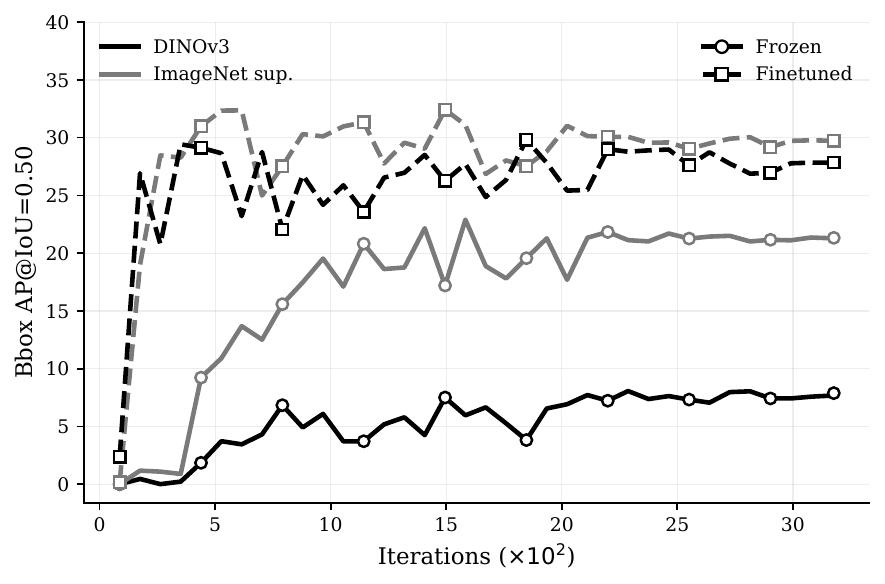}
   \caption{Learning curves for bounding box AP on the GDXray validation set using Faster R-CNN with ConvNeXt-T bakbones pretrained with supervised ImageNet classification or DINOv3 distillation.}
   \label{fig:gdxray_results}
\end{figure}

Table~\ref{tab:comparison_pretraining} reports the final validation performance of all compared backbones and pretraining strategies, and \cref{fig:sem_seg_results,fig:rareplanes_results,fig:gdxray_results} show the corresponding learning curves under frozen and fully finetuned regimes. Two main observations emerge. First, the effect of pretraining is strongly conditioned by the target domain: on RGB datasets, DINOv3 distillation does not provide a clear advantage in the frozen setting and only becomes consistently beneficial after full finetuning, whereas on X-ray imagery it remains inferior to supervised ImageNet initialization in both regimes. Second, the backbone family itself matters: independently of the pretraining strategy, ConvNeXt consistently improves over the conventional ResNet-50 baseline once fully finetuned, indicating that stronger convolutional architectures already provide a solid transfer benefit for industrial visual inspection. More specifically, DINOv3-distilled ConvNeXt yields the best results on all RGB tasks after full adaptation, while ImageNet-pretrained ConvNeXt remains the strongest choice under X-ray modality shift. These results suggest that large-scale self-supervised pretraining on web-data can provide a stronger initialization for RGB industrial inspection, but that its transferability degrades substantially when the target modality departs too far from natural-image statistics.

\subsection{Semantic Segmentation}
The semantic segmentation results on Severstal and Rubber Rings reveal a clear distinction between the probing and full-finetuning regimes. When the backbone is frozen, the gap between the two pretraining strategies remains limited. On Severstal, ConvNeXt-T pretrained with supervised ImageNet classification and with DINOv3 distillation achieve very similar performance (62.04 vs. 62.40 mIoU), suggesting that, in this setting, DINOv3 does not provide a strong advantage over conventional pretraining when the backbone is used only as a fixed feature extractor. A similar trend is observed on Rubber Rings, where ImageNet pretraining is even slightly stronger than DINOv3 in the frozen regime (73.25 vs. 72.32 mIoU). Overall, these results indicate that for industrial RGB semantic segmentation, the superiority of DINOv3 pretraining is not immediately visible under a linear-probing-style protocol, in contrast to the large gains often reported on natural-image benchmarks.

A different behavior emerges once the backbone is fully finetuned. In this regime, DINOv3-distilled ConvNeXt-T achieves the best performance on both datasets, outperforming both the conventional ResNet-50 baseline ($+10.59$ mIoU) and the ImageNet-pretrained ConvNeXt-T ($+10.29$ mIoU). This result is particularly notable because the ImageNet-pretrained ConvNeXt-T does not improve over its frozen counterpart on either dataset, whereas DINOv3 pretraining yields a clear gain after adaptation. This suggests that, although DINOv3 features do not offer a strong advantage as frozen descriptors for industrial surface defects, they provide a more favorable initialization for downstream finetuning.

The learning curves further support this interpretation. As shown in Fig.~\ref{fig:sem_seg_results}, the frozen-backbone experiments exhibit similar optimization behavior for both pretraining strategies, consistent with their close final mIoU values. In contrast, under full finetuning, DINOv3 initialization leads to faster convergence and higher final performance, indicating that the dense visual features learned during pretraining facilitate downstream adaptation.

These results indicate that, for industrial RGB semantic segmentation, DINOv3 acts primarily as a superior finetuning initialization rather than as a stronger frozen feature extractor.

\subsection{RGB Instance Segmentation and Localization}
The results on RarePlanes show a behavior similar to semantic segmentation. When the backbone is frozen, supervised ImageNet pretraining performs better than DINOv3 distillation. However, this trend reverses under full finetuning, where DINOv3-distilled ConvNeXt-T achieves the best result (\(84.50\) mask mAP), outperforming both the ImageNet-pretrained ConvNeXt-T (\(82.88\)) and the ResNet-50 baseline (\(78.39\)). The learning curves confirm this behavior: ImageNet pretraining remains slightly stronger in the frozen setting, whereas DINOv3 leads to better convergence and higher final performance once the backbone is fully adapted.

A plausible explanation is that DINO self-distillation primarily learn a global semantic invariance, while Mask R-CNN relies more heavily on instance-level localization and bounding-box regression. Prior work has shown that detection benefits more when pretraining is aligned with multi-object data and regression-aware objectives \cite{liAlignDetAligningPretraining2023}, and that DINO-like joint-embedding methods are less robust on non-object-centric datasets than masked-image-modeling approaches \cite{el-noubyAreLargescaleDatasets2021}. This may explain why DINOv3 provides only limited gains in frozen transfer for detection, while still offering a stronger initialization for finetuning. We should nevertheless keep in mind that the student backbone remains a convolutional architecture, whose transfer behavior is still shaped by CNN-specific inductive biases such as locality, weight sharing, downsampling, and multichanneling \cite{wangTheoreticalAnalysisInductive2023}. This suggests that the limited gains of DINOv3 in frozen transfer may also reflect a partial mismatch between the distilled representation from a \textit{Vision Transformer} and the convolution-friendly priors of the ConvNeXt student that remain important for localization-oriented transfer, even if the same initialization becomes highly effective once the backbone is fully finetuned.

\subsection{X-ray Defect Detection}
The results on GDXray Castings reveal a markedly different behavior from the RGB datasets and highlight the difficulty of transferring web-pretrained representations under modality shift. When the backbone is frozen, the performance gap is already large: ImageNet-pretrained ConvNeXt-T reaches 21.32 box mAP@50, whereas the DINOv3-distilled counterpart drops to only 7.88. This gap remains after full finetuning, where ImageNet pretraining still yields the best result (29.74) compared with DINOv3 distillation (27.84).

These results suggest that the representations learned through DINOv3 distillation on a ConvNeXt do not transfer effectively to X-ray imagery, even after downstream adaptation. Unlike RGB surface inspection, X-ray images differ strongly from natural-image web data in terms of appearance statistics, texture structure, and object formation, which likely weakens the usefulness of the semantic priors encoded by self-supervised pretraining. In contrast, supervised ImageNet initialization appears to provide a more stable starting point for this detection task. Overall, the GDXray results indicate that the benefits of DINOv3 do not extend to strong modality shifts, and that conventional supervised pretraining remains more reliable in this setting.


%% file: sec/6_discussions.tex
\section{Discussions}
\label{sec:discussions}
Our study shows that the behavior of DINOv3-pretrained ConvNet backbones on industrial visual inspection tasks depends strongly on both the downstream task and the target modality. On RGB datasets, compared with conventional supervised ImageNet pretraining, DINOv3 pretrained features does not provide a clear advantage, but provides the strongest initialization once the backbone is fully finetuned, leading to faster convergence and better final performance for semantic segmentation and RGB instance-level localization. In contrast, this advantage disappears on X-ray defect detection, where supervised ImageNet pretraining remains superior in both frozen and fully finetuned settings. Overall, these observations suggest that the benefit of modern vision foundation models in industrial inspection lies less in direct universal transfer than in their ability to provide strong finetuning priors when the target data remain sufficiently close to natural-image statistics.

These observations help clarify three practical questions for industrial transfer learning with modern vision foundation models: 
\paragraph{Direct transferability of DINOv3 representations.}
Only to a limited extent. In our experiments, DINOv3 does not consistently outperform conventional ImageNet pretraining when the backbone is frozen, and it fails clearly under X-ray modality shift. This indicates that web-scale pretrained features are not universally aligned with industrial imagery. However, the gains observed after full finetuning on RGB tasks show that these representations still provide a valuable initialization when sufficient adaptation is allowed.

\paragraph{Differences from conventional ImageNet transfer.}
Our results suggest that the two strategies transfer differently. ImageNet pretraining remains a robust baseline, especially in frozen settings and under strong modality shift. By contrast, DINOv3 offers limited benefit as a fixed feature extractor, but a stronger starting point for downstream adaptation on RGB tasks. In this sense, DINOv3 appears less effective as a universal frozen descriptor, but more effective as a finetuning initialization.

\paragraph{Toward industrial-specific self-supervised pretraining.}
Self-supervised learning has opened the way to vision foundation models with broadly transferable representations, and dense-oriented approaches such as DINOv3 may become a new standard for transfer learning. However, the notion of universal visual representation remains constrained by the distribution of the data used during pretraining, which is still dominated by natural-image web data. For industrial inspection, where visual patterns are often shaped by specific materials, acquisition setups, and non-standard modalities, an important next step is to move toward industrial-specific self-supervised pretraining. Such a direction could enable the development of foundation models that are not only transferable, but also intrinsically better matched to real industrial environments.

%% file: sec/7_acknowledgments.tex
\section*{Acknowledgements} This work made in FACTOLAB (commun laboratory UCA, CNRS, Michelin) and International Research Center "Innovation Transportation and Production Systems" of the I-SITE CAP 20-25 frameworks was supported by Michelin Tyres Manufacturer.